\documentclass[runningheads]{llncs}

\usepackage{eccv}

\usepackage{eccvabbrv}

\usepackage{graphicx}
\usepackage{booktabs}
\usepackage{bbding}

\usepackage[accsupp]{axessibility}  

\usepackage{hyperref}

\usepackage{orcidlink}

\begin{document}

\title{MM2Latent: Text-to-facial image generation and editing in GANs with multimodal assistance} 


\author{Debin Meng\inst{1}\orcidlink{0009-0002-0430-0660} \and
        Christos Tzelepis\inst{1,2}\orcidlink{0000-0002-2036-9089} \and
        Ioannis Patras\inst{1}\orcidlink{0000-0003-3913-4738} \and
        Georgios Tzimiropoulos\inst{1,2}\orcidlink{0000-1111-2222-3333}}

\authorrunning{D.~Meng et al.}

\institute{Queen Mary University of London, London E1 4NS, UK \and
Samsung AI Center, Cambridge, UK \\
\email{\{debin.meng, c.tzelepis, i.patras, g.tzimiropoulos\}@qmul.ac.uk}}

\maketitle

\begin{figure}[h]
   \begin{center}
   \includegraphics[width=0.99\linewidth]{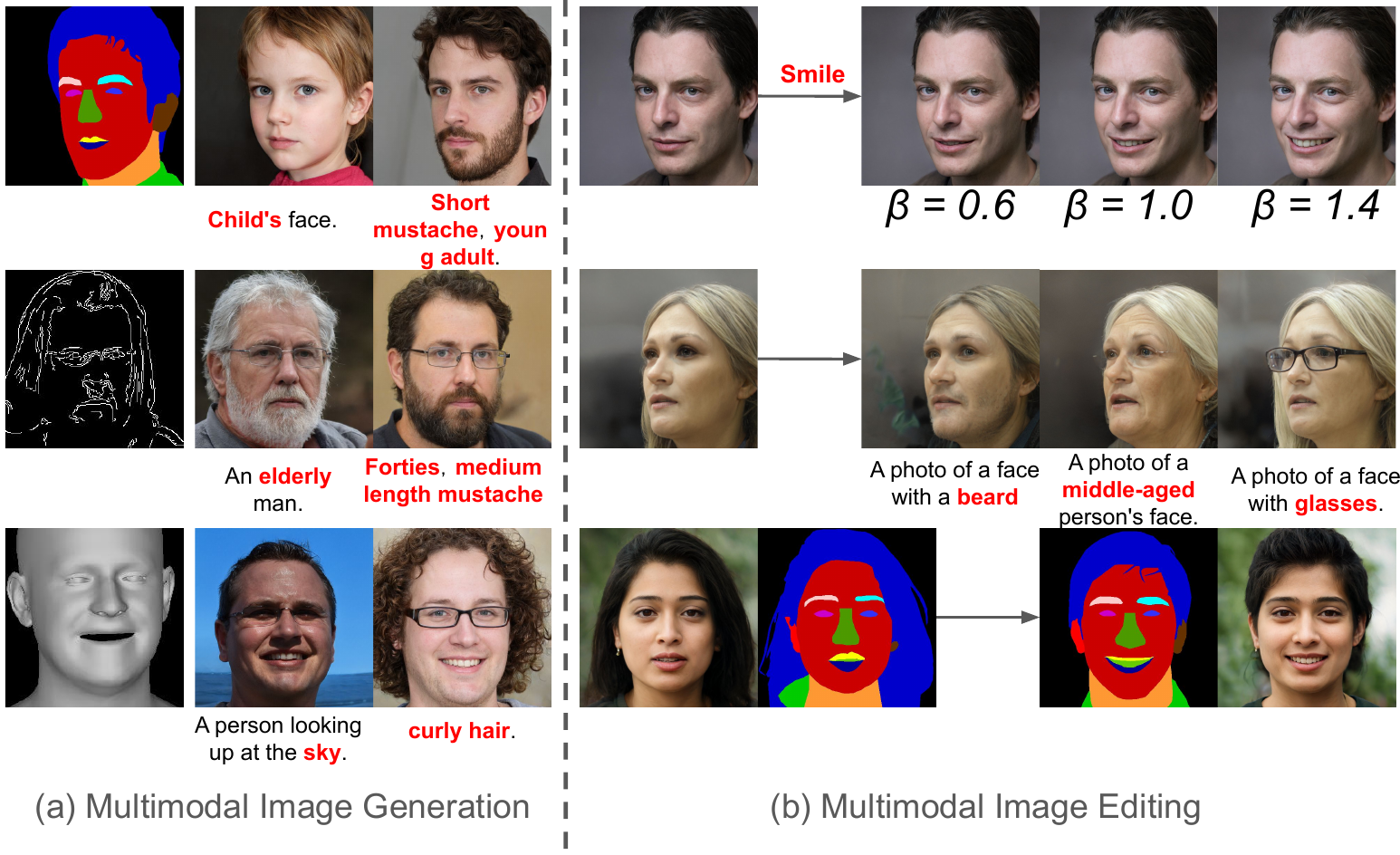}
   \end{center}
   \vspace{-1em}
   \caption{We propose MM2Latent, a versatile framework for multimodal image generation and editing using facial segmentation masks, sketches, and 3DMM parameters.} 
   \vspace{-1em}
   \label{fig:teaser}
\end{figure}

\begin{abstract}
Generating human portraits is a hot topic in the image generation area, e.g. mask-to-face generation and text-to-face generation. However, these unimodal generation methods lack controllability in image generation. Controllability can be enhanced by exploring the advantages and complementarities of various modalities. For instance, we can utilize the advantages of text in controlling diverse attributes and masks in controlling spatial locations. Current state-of-the-art methods in multimodal generation face limitations due to their reliance on extensive hyperparameters, manual operations during the inference stage, substantial computational demands during training and inference, or inability to edit real images. In this paper, we propose a practical framework —~MM2Latent~—~for multimodal image generation and editing. We use StyleGAN2 as our image generator, FaRL for text encoding, and train an autoencoders for spatial modalities like mask, sketch and 3DMM. We propose a strategy that involves training a mapping network to map the multimodal input into the w latent space of StyleGAN. The proposed framework 1) eliminates hyperparameters and manual operations in the inference stage, 2) ensures fast inference speeds, and 3) enables the editing of real images. Extensive experiments demonstrate that our method exhibits superior performance in multimodal image generation, surpassing recent GAN- and diffusion-based methods. Also, it proves effective in multimodal image editing and is faster than GAN- and diffusion-based methods. We make the code publicly available at: \url{https://github.com/Open-Debin/MM2Latent}.
\end{abstract}

\section{Introduction}\label{sec:intro}

    Generating human portraits~\cite{chen2019ftgan, nasir2019text2facegan, sun2021multi} has emerged as a prominent sub-task in the conjunction of generative learning, computer vision, and multimedia~\cite{hou2022textface, patashnik2021styleclip, ramesh2021zero, xia2021tedigan}, drawing significant attention from both academia and industry due to its potential applications in art, design, entertainment, and advertising. Recently, there have been many advancements in image generation techniques, such as generative adversarial networks (GANs)~\cite{brock2018large, creswell2018generative, goodfellow2020generative, karras2019style} and diffusion models~\cite{dhariwal2021diffusion, ho2020denoising,song2020score, sohl2015deep}, which have enabled the generation of synthetic images of unprecedented quality and diversity. 

    In addition to improving the generation quality of fundamental generative models (e.g., GANs and Diffusion Models), controllability of generation has emerged as an open and challenging problem towards meeting users' diverse requirements for image synthesis and editing. An example of such conditioning signals is natural language -- i.e., text descriptions for controllable generation (i.e., Text-to-Image generation~\cite{nasir2019text2facegan,chen2019ftgan,peng2021knowledge,sun2021multi, tao2020deep,wang2021faces,zhu2019dm}), which aims to close the gap between semantic descriptions and visual content, allowing for the creation of facial images that faithfully represent the described attributes and characteristics.

    While natural language offers flexibility and versatility, its inherent ambiguity poses notable challenges in accurately controlling spatial generation. For instance, it is difficult to accurately describe the shape of face using natural language alone. In contrast, visual signals offer more precise spatial information compared to language. Therefore, many studies have  utilized visual modalities for more accurate and controllable image generation, such as facial segmentation mask~\cite{chen2022sofgan,chen2017photographic,liu2019learning,mirza2014conditional,qi2018semi,sushko2020you,wang2018high, wang2021image, zhu2020sean, zhu2020semantically}, sketches~\cite{richardson2021encoding,chen2020deepfacedrawing,wang2018high}, and 3D Morphable Models (3DMM)~\cite{tewari2020stylerig, bounareli2023stylemask, bounareli2024one, bounareli2023hyperreenact, bounareli2024diffusionact}. Compared to language, segmentation masks can define the position and shape of face more precisely. However, visual spatial information lacks controllability in semantic attributes, such as hair color, age, and gender.

    The complementary advantages of visual and language modalities enable them to compensate for each other's limitations. For instance, we can utilize the advantages of text in controlling diverse attributes and masks in controlling spatial locations. Recent works in multimodal image generation include mainly GAN-based~\cite{xia2021tedigan, du2023pixelface+} or Diffusion-based~\cite{nair2023unite, huang2023collaborative, zhang2023adding} methods. However, these methods are significantly limited by their reliance on manual tuning of many hyper-parameters and/or manual operations~\cite{xia2021tedigan} during the inference stage or have significant computational demands both in training and inference~\cite{nair2023unite, huang2023collaborative, zhang2023adding}. Du et al.~\cite{du2023pixelface+} provide a framework for multimodal image generation and editing but the proposed method is applied only on synthetic, not real images.

    In this paper, we propose~MM2Latent, a novel framework for multimodal image generation and editing. Compared to existing approaches, our method: 1) does not require manual tuning of hyper-parameters or manual operations during the inference stage, 2) ensures fast inference speeds, and 3) enables the editing of real images. The proposed MM2Latent uses StyleGAN2 as our image generator, FaRL~\cite{zheng2022general} for text encoding, and autoencoders for spatial modalities like mask, sketch, and 3DMM. We propose a strategy that involves training a mapping network to map the multimodal input into the $\mathcal{W}$ latent space of StyleGAN. Specifically, the proposed MappingNetwork is trained on image embeddings but accepts text embeddings at the inference stage due to the visual language alignment of FaRL~\cite{zheng2022general}. To increase its generalization ability, we generate pseudo text embeddings during training. The MappingNetwork can predict image editing directions in the latent space of StyleGAN. We achieve multimodal facial editing by applying the editing direction on faces inverted by a GAN inversion method (e4e~\cite{tov2021designing}).

    Extensive experimental evaluations demonstrate that proposed MM2Latent outperforms current state-of-the-art methods in terms of multimodal consistency, image quality, and inference speed. The main contributions of our work are summarized as follows:
    \begin{itemize}
        \item We propose MM2Latent, a novel multimodal StyleGAN-based synthesis method for controllable facial image generation using text combined with masks, sketches, or 3DMM.
        \item MM2Latent allows for interactive face editing of real images. It provides multiple editing controls, such as text, mask/sketch/3DMM-guided editing, offering flexible control over facial semantic and spatial attributes.
        \item Extensive quantitative and qualitative experiments demonstrate the advancement of our framework in achieving better multimodal consistency, higher image quality, and faster inference speed. 
    \end{itemize}

\section{Related Work}\label{sec:related_work}

    \subsection{Image Generation}

        Image synthesis is an important task in the conjunction of generative learning, computer vision, and multimedia~\cite{Afifi2020HistoGANCC, He2021EigenGANLE, Liang2021HighResolutionPI, Lin2021AnycostGF, Liu2021DivCoDC}. Generative Adversarial Networks (GANs)~\cite{Goodfellow2014GenerativeAN, karras2019style} have played a remarkable role in image synthesis due to their unprecedented ability in generating realistic and aesthetically pleasing images, often indistinguishable from real ones, paving the way towards application such as face reenactment~\cite{bounareli2023hyperreenact, bounareli2023stylemask, bounareli2024one, bounareli2024diffusionact}, image editing~\cite{tzelepis2021warpedganspace, oldfield2023panda, oldfield2024bilinear, tzelepis2022contraclip, d2024improving}, and face anonymization~\cite{barattin2023attribute}.
        
        More recent advancements in generative learning include Diffusion Probabilistic Models (DPMs)~\cite{ho2020denoising} that despite their remarkable ability to produce realistic and diverse synthetic images, their application scope is limited by the vast compute power and data they require for training and their slow and less controllable inference process. To solve these limitation, Denoising diffusion implicit models (DDIM)~\cite{Song2020DenoisingDI} were proposed for faster and deterministic inference, whilst Latent Diffusion Models (LDMs)~\cite{Rombach2021HighResolutionIS} proposed to operate the diffusion process in a lower-dimensional latent space, resulting in lower training and inference costs.

    \subsection{Conditional Face Generation}
        
        Conditional face generation aims at generating high-quality face images conditioned on a given signal. Common conditioning signals include text prompts~\cite{Peng2022LearningDP, Peng2022TowardsOT, Sun2021MulticaptionTS, oldfield2023parts}, segmentation masks~\cite{CelebAMask-HQ}, and 3D Morphable Model (3DMM) parameters~\cite{tewari2020stylerig, bounareli2024one, bounareli2023hyperreenact}. Such methods typically incorporate unimodal conditions, and are thus limited by the limitations of each modality. For instance, the inherent ambiguity of natural language poses certain challenges in accurately controlling spatial features. Yet, visual spatial information, such as segmentation masks that can accurately condition spatial information, lack controllability in semantic attributes, such as hair color, age, and gender.

        To address these limitations of unimodal methods, multimodal face generation methods aim to combine the complementary advantages of multiple modalities to create a highly controllable generation model. Composable Diffusion~\cite{Liu2022CompositionalVG} has demonstrated the complementary abilities of diffusion models in the latent noise space. ControlNet~\cite{Zhang2023AddingCC} fine-tunes the pretrained Latent Diffusion Models (LDMs)~\cite{Rombach2021HighResolutionIS} to enable diffusion models to accept inputs from multiple modalities. TediGAN~\cite{xia2021tedigan} is a StyleGAN-based face synthesis and manipulation method that performs style mixing in the StyleGAN latent space to achieve multimodal generation. PixelFace+~\cite{du2023pixelface+} incorporates pixel synthesis~\cite{he2022pixelfolder} and CLIP~\cite{Radford2021LearningTV}. Collaborative Diffusion~\cite{huang2023collaborative} and UniteConquer~\cite{nair2023unite} extend the compositional diffusion model, by learning models to weight and fuse latent noise from multiple diffusion models, or by involving classifier-free guidance in multimodal image generation, respectively.

    \subsection{Face Manipulation}
        
        For real face manipulation (i.e., editing of real images), existing works typically involve the inversion of the real images onto the latent space of a generative model (e.g., the $\mathcal{W}$ space of StyleGAN2~\cite{karras2020analyzing}) and the manipulation of the respective latent codes according to certain criteria (e.g., towards specific facial attributes or head pose). Imagic~\cite{Kawar2022ImagicTR} is a diffusion-based method that fine-tunes both the text embedding and the generative model for each image editing task, resulting in significant time and memory costs. Null-text inversion~\cite{mokady2023null} and Prompt Tuning Inversion~\cite{dong2023prompt} only fine-tune their unconditional embeddings (i.e., null text embedding), leading to more memory-efficient generation compared to Imagic. However, these methods remain notably slow for real-world applications. In contrast to diffusion-based methods, GAN-based inversion methods generally either (i) directly optimize the latent space to minimize the error for the given image, or (ii) train an encoder to map the given image to the latent space, or (iii) use a hybrid approach combining both. Typically, methods that perform optimization are superior in achieving higher reconstruction quality, but are slower than encoder mapping methods. For image editing in StyleGAN, the $\mathcal{W}$ and $\mathcal{W}+$ latent spaces are commonly used. $\mathcal{W}$ is typically the preferred latent space for image editing, while $\mathcal{W}_+$ for image reconstruction~\cite{tov2021designing} -- e4e~\cite{tov2021designing} is a standard GAN inversion method for StyleGAN2~\cite{karras2020analyzing} that leads to a good trade-off between faithful reconstruction and editability and has been used extensively in image editing tasks~\cite{Patashnik2021StyleCLIPTM, Pinkney2022clip2latentTD, bounareli2023hyperreenact}.

\section{Proposed Method}\label{sec:proposed_method}

    \begin{figure*}[htp]
        \centering
        \includegraphics[width=1.01\textwidth]{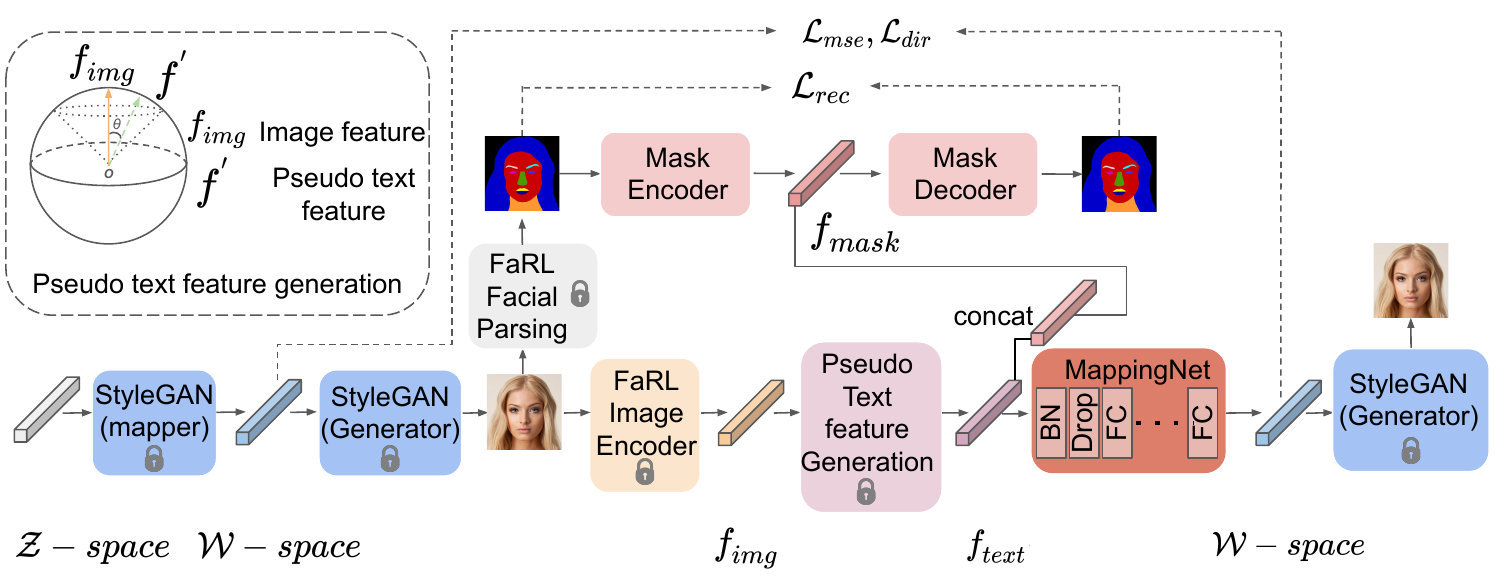}
        \caption{Overview of the proposed MM2Latent's training process. First, the mask autoencoder is trained followed by the training of the MappingNet while keeping the other modules fixed  -- note that we show only the mask modality for brevity.}
        \label{fig:main_pipeline}
    \end{figure*}

    \subsection{Main components of MM2Latent}
        Multimodal image generation consists in generating images from various input modalities. If we consider text and mask modality, the general method can be defined as follows:
        \begin{equation}
            w=\mathbf{Net}(\mathbf{F}_{mask}(x_m),\:\mathbf{F}_{text}(x_t)),\:I=\mathbf{G}(w),
        \end{equation}
        where $x_t$, $x_m$, $\mathbf{F}_{text}(\cdot)$, and $\mathbf{F}_{mask}(\cdot)$ denote the text input, the mask input, the text and mask the encoder for text, and the encoder for mask, respectively. $\mathbf{Net}(\cdot)$ denotes the multimodal fusion module, which predicts image latent embedding $w$ by fusing the multimodal input from $\mathbf{F}_{text}(\cdot)$ and $\mathbf{F}_{mask}(\cdot)$. Finally, the image latent embeddings $w$ are fed to a generator $\mathbf{G}(\cdot)$ to produce the output image. The challenge is designing the multimodal fusion module $\mathbf{Net}(\cdot)$, conditional encoder $\mathbf{F}_{text}(\cdot)$ and $\mathbf{F}_{mask}(\cdot)$. 

        \subsubsection{The designing of multimodal fusion}
            We propose to use an MLP stack by multi-fully connected layers (FC) to map multimodal features to the image latent space (see the first row in Tab. ~\ref{tab:ablation_componets}).

            \begin{figure}[htp]
                \centering
                \includegraphics[width=1\textwidth]{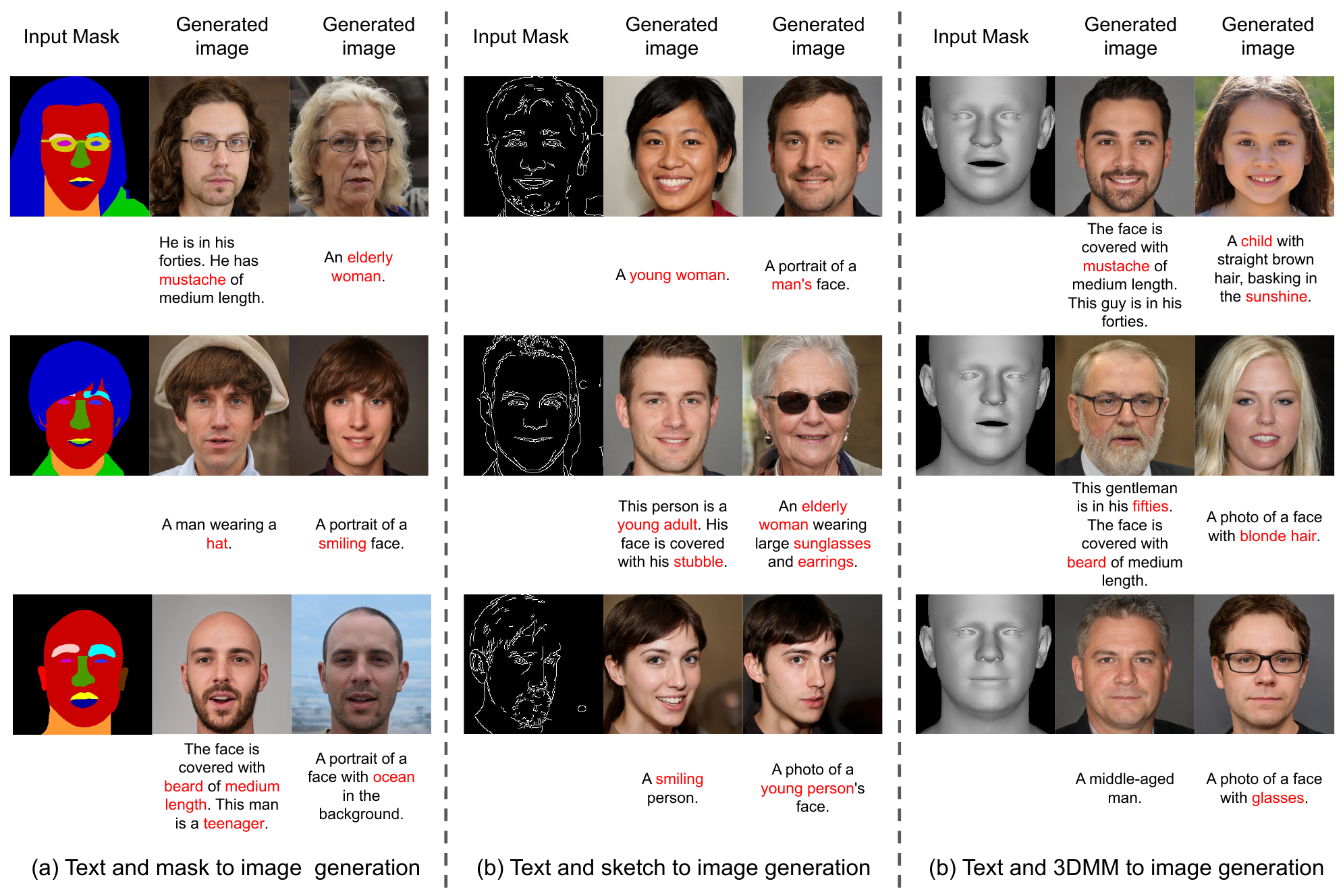}
                \caption{Multimodal image generation. Each generated image is accompanied by a textual description below it and a spatial mask, sketch, or 3DMM to its left. }
                \label{fig:mask_sketch_3dmm_synthetic}
            \end{figure}

            Firstly, without the component of Pseudo Text embedding generation (see Fig.~\ref{fig:main_pipeline}), the input of MappingNet is the $f_{mask}$ and $f_{img}$, which represent the mask embeddings and text embeddings respectively. The $f_{mask}$ and $f_{img}$ come from the sample, so they are highly correlated. However, in the inference stage, the text prompts are not always highly correlated with the mask (e.g. the user may expect to generate people of different genders and attributes based on the same mask.). In order to simulate the situation in the inference phase (where one mask may be combined with diverse text), during the training phase, we involve the component of pseudo text embedding generation inspired by~\cite{Pinkney2022clip2latentTD, Zhou2021TowardsLT}. This component generates the pseudo text embeddings $f^{'}_{text}$ from $f_{img}$. The $f^{'}_{text}$ is concatenated with $f_{mask}$ as the input of the MappingNet. This component has two purposes in our framework, 1. the generated $f_{text}$ simulates the situation in the inference phase (where one mask may be combined with diverse text), thereby increasing the generalizability of the MappingNet. 2. It plays the role of data augmentation because one $f_{img}$ can generate multiple $f_{text}$, which enriches the training dataset. The formulation of the Pseudo Text embedding generation is defined as:
            \begin{equation}
                f^\prime_{text}=\frac{y}{\|y\|_2}, \quad y=f_{img}+ \frac{\varepsilon}{\|\varepsilon\|_2},
            \end{equation}
            where $\varepsilon\in\mathcal{N}(0,I)$ is a Gaussian noise vector of the same dimension as $f_{img}$. 

        \subsubsection{The encoding of text}
            We adopt the FaRL~\cite{zheng2022general} text encoder. FaRL is a visual-language joint model, trained on 20 million facial image-text pairs. FaRL has already demonstrated its excellent performance in facial attribute encoding and has been adopted in previous SoTA multimodal image generation work~\cite{nair2023unite}.

        \subsubsection{The encoding of mask}
            For the mask encoder, we train a mask autoencoder from scratch and use its encoder part $\mathbf{F}_{mask}(\cdot)$ in the forward path of our multimodal generation pipeline. The mask autoencoder is defined as follows:
            \begin{equation} 
                f_{mask} = \mathbf{F}_{mask}(x_m), ~\hat{x}_m = \mathbf{D}_{mask}(f_{mask})
            \end{equation}
            Here $\mathbf{D}_{mask}(\cdot)$ is the mask decoder, the predicted $\hat{x}_m$ should have reconstructed the input $x_m$. MSE loss is adopted for training this autoencoder to ensure each pixel of the input mask $x_m$ has been well reconstructed:
            \begin{equation} 
            \mathcal{L}_{mse} = \frac{1}{n}\sum_{i=0}^{n}\sum_{j=0}^{d}(x_{ij} - \hat{x}_{ij})^2 
            \end{equation}
            Our mask autoencoder only stacks basic convolutional, pooling, and non-linear activation layers. Please refer to the supplementary material for more implementation details.

        \subsubsection{The encoding of sketch}
            Similarly to the mask modality, we train an autoencoder from scratch. The primary difference is in the training losses, given that sketch images contain only two pixel values (0 and 255) representing the background and the sketch, respectively. We treat this as a binary classification task, employing binary cross-entropy loss to train the encoder:
            \begin{equation} 
                \mathcal{L}_{sketch} = -\frac{1}{n}\sum_{i=0}^{n}\sum_{j=0}^{d}(x_{ij}log\hat{x}_{ij} + (1-x_{ij})log(1-\hat{x}_{ij})) 
            \end{equation}

        \subsubsection{The encoding of 3DMM}
            We adopt a 3DMM encoder from DECA~\cite{Feng2020LearningAA}, a state-of-the-art open-source 3D reconstruction framework. DECA utilizes an autoencoder architecture based on 3DMM to convert RGB images into 159-dimensional 3DMM parameters. These parameters include 100 for facial shape, 50 for facial expression, and 9 for facial and camera pose. In our approach, we use these 3DMM parameters as our 3DMM conditional embeddings. 

        \subsubsection{The image generator}
            We adopted styleGAN~\cite{karras2019style} as our generator, which has semantically rich and disentangled $w$-latent space and has high quality in facial image generation. Our MappingNet predicts the multimodal $f_{text}$ and $f_{mask}$ to $w$-latent, then the realistic image is generated from the generator:
            \begin{equation}
                w = \mathbf{MappingNet}(f_{text}, f_{mask}), ~~I = \mathbf{G}(w).
            \end{equation}

        \subsection{Training losses}
            The proposed MappingNet's goal is to predict the latent code $\hat{w}$ that will drive the generation of the desired face. For doing so, we propose to optimise the following loss function:
            \begin{equation} \label{eq:mapping_total_loss}
                \mathcal{L}_{total} = \frac{1}{n} \sum_{i=1}^{n} \mathcal{L}_{abs}(w_i,\hat{w} _i) + \lambda \cdot \frac{1}{n}\sum_{i=1}^{n}\mathcal{L}_{dir}(w_i,\hat{w} _i),
            \end{equation}
            where $\mathcal{L}_{abs}$ denotes the absolute value loss given as $\mathcal{L}_{abs}(x,y) = \frac{1}{d}\sum_{i=0}^{d}(x^i - y^i)^2$ and $\mathcal{L}_{dir}$ denotes the direction loss given as $\mathcal{L}_{dir}(x,y) = 1 - \frac{x\cdot y}{|x|\cdot |y|}$, where $d$ denotes the the dimensionality of the $\mathcal{W}$ space and $n$ the batch size. $\lambda$ is the weighting hyper-parameter empirically set to $\lambda=10$.

        \subsection{Training the whole framework}
            Training multimodal image generation models typically requires a large-scale dataset comprising image-text pairs. However, labelling text descriptions generally is both time-intensive and costly. Our framework leverages the visual and language alignment in FaRL to avoid being limited to the image-text pairs dataset. Specifically, our work requires the knowledge of the ground truth $w$ of the input $f_{text}$ and $f_{mask}$. By sampling the $\mathcal{Z}$-space of the StyleGAN, images and its ground truth $w$ are generated. Then a third-party facial parsing method~\cite{zheng2022general} is applied to the images to generate facial masks for training (facial sketch is generated by OpenCV~\cite{opencv_library}). We use FaRL image encoder to get the image embeddings $f_{img}$. We use mask/sketch encoder to extract $f_{mask}/f_{sketch}$. After generating the $f^{'}_{text}$ from $f_{img}$ by the Pesudo text embedding generator, we now have the input $f^{'}_{text}$ and $f_{mask}$ and their ground truth $w$ for training the framework.

        \subsection{Inference}
            
            \subsubsection{Multimodal Image Generation}
                Although our framework is trained on the Pesudo text embeddings generated from image embeddings of FaRL, during the inference stage, we can directly use the real text embeddings of FaRL  for multimodal image generation. Since the image and text space of FaRL have been aligned, we leverage its alignment attributes in the inference stage to avoid the need for any text labelled data for training our multimodal MappingNet.

            \subsubsection{Multimodal Image Editing}
                Once the original image $I_{src}$ is encoded into the $\mathcal{W}+$ space of StyleGAN as $w_{src}$, we can edit it by navigating it along the semantic meaningful $w_{dir}$ direction. For example, if the $w_{dir}$ direction can make face older, the edited $w_{edit}$ from $w_{edit} = w_{src} + \beta \cdot w_{dir}$ can be generated an older face than the $w_{src}$. Our task is to find the $w_{dir}$ in a multimodal way.

                For multimodal text editing, we rely on pivotal text and target text, such as ``A photo of a person'' and ``A photo of a person with a beard'', respectively. Then we get the $f_{piv}$ and $f_{tar}$. For real image $I_{src}$ and its latent embedding $w_{src}$ in StyleGAN and $f_{mask}$, we can find the $w_{dir}$ as follows:
                \begin{equation} 
                    w_{dir} = Net(f_{tar}, f_{mask}) - Net(f_{piv}, f_{mask}),
                \end{equation}
                where $Net(\cdot)$ is the mapping network. Similarly, we can get the $w_{dir}$ from multimodal mask editing, where we use the image embedding $f_{img}$ and mask embeddings $f_{mask\_tar}$ and $f_{mask\_piv}$ for this task:
                \begin{equation} 
                    w_{dir} = Net(f_{img}, f_{mask\_tar}) - Net(f_{img}, f_{mask\_piv}).
                \end{equation}

\section{Experiments}\label{sec:experiments}
    In this section, we quantitatively evaluate our method on multimodal consistency (including text and mask consistency) and image quality. We condacted extensive experiments on FFHQ~\cite{karras2019style} and the multimodal text-to-image generation benchmarks CelebAHQ-Mask~\cite{CelebAMask-HQ}~/~Dialog~\cite{Jiang2021TalktoEditFF}. Our method is compared with open-source state-of-the-art techniques in multimodal face generation, namely TediGAN~\cite{xia2021tedigan}, Composal~\cite{Liu2022CompositionalVG}, UniteConquer~\cite{nair2023unite}, and Collaborative Diffusion~\cite{huang2023collaborative}.

    \subsection{Experimental Setup}
        \textbf{Dataset} The evaluation utilizes mask and text pairs from CelebAHQ-Mask, with corresponding textual descriptions available in CelebA-Dialog. CelebAHQ-Mask~\cite{CelebAMask-HQ} features manually annotated segmentation masks for 30000 images from CelebA-HQ~\cite{karras2017progressive}. Each mask categorizes up to 19 classes, including primary facial components such as hair, skin, eyes, and nose, as well as accessories like eyeglasses and clothing. CelebA-Dialog~\cite{Jiang2021TalktoEditFF} provides fine-grained natural language descriptions for the images in CelebA-HQ. FFHQ~\cite{karras2019style} comprises 70000 high-resolution and high-quality real facial images. We use this dataset to evaluate the image quality by computing the CMMD~\cite{jayasumana2024rethinking} distance between the generated images and whole set of real high-quality images from FFHQ.

    \subsection{Evaluation Metrics}
        In multimodal face image generation, we assess the consistency between the generated image and the multimodal input signals. Specifically, we evaluate text-to-image consistency using the CLIP Score and mask-to-image consistency using Mask Accuracy. Also, we assess image quality using the CMMD metric~\cite{jayasumana2024rethinking}.
 
        \noindent \textbf{CLIP Score} CLIP~\cite{Radford2021LearningTV} is a large-scale vision-language model that employs separate encoders for images and texts to project them into an aligned feature space. The CLIP score is calculated as the cosine similarity between the normalized embeddings of an image and text. Generally, a higher score indicates greater consistency between the generated image and the corresponding text caption.

        \noindent \textbf{Mask Accuracy} For each generated image, we predict the segmentation mask using the face parsing network from CelebAMask-HQ \cite{CelebAMask-HQ}. Mask accuracy is determined by the pixel-wise accuracy compared to the ground-truth segmentation. Higher average accuracy indicates better consistency between the output image and its corresponding segmentation mask.

        \begin{figure}[htp]
          \centering
          \includegraphics[width=1\textwidth]{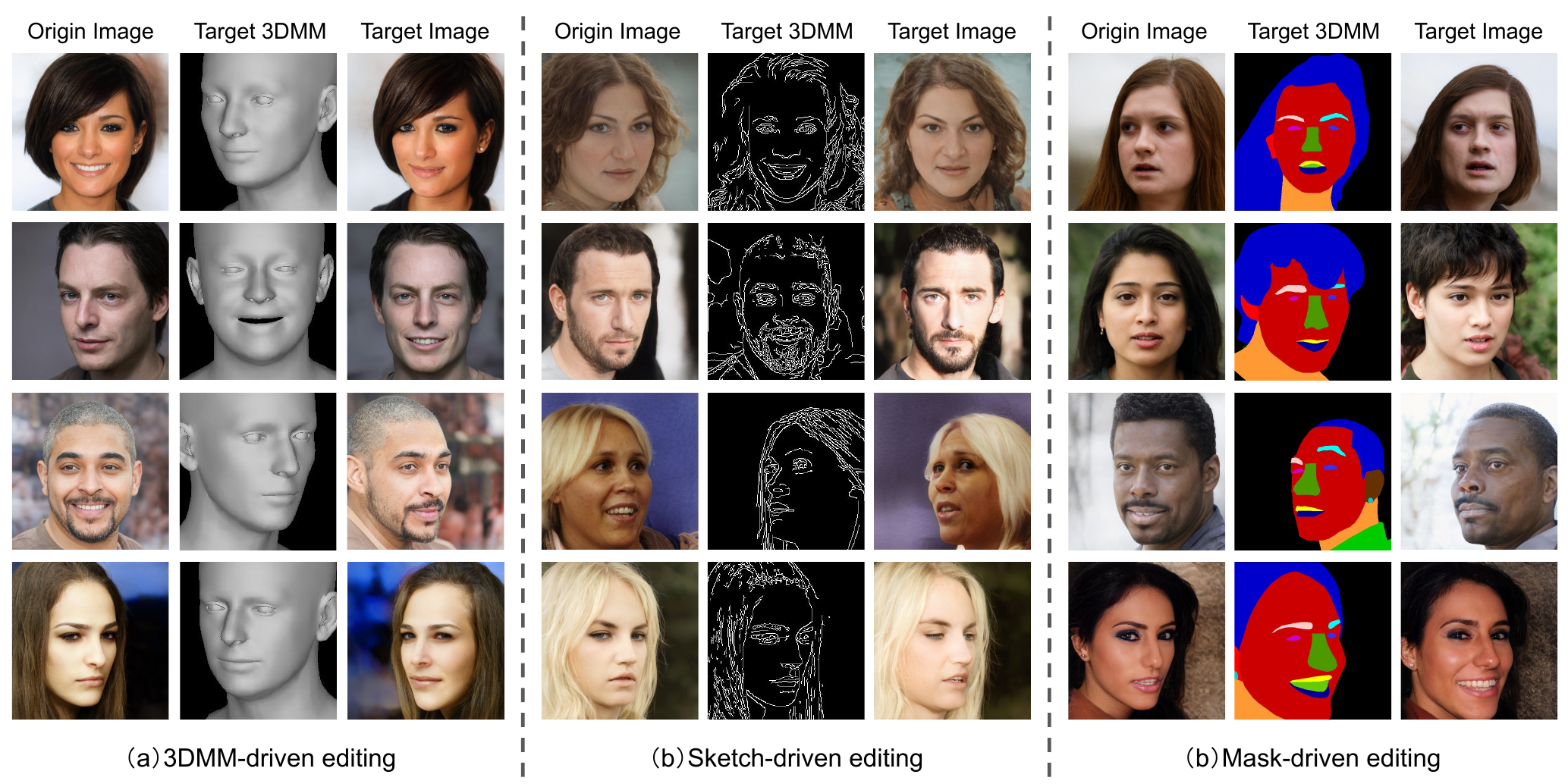}
          \caption{Multimodal spatial editing. we focus on modifying the shape of the original image according to targeted spatial information, while preserving its inherent attributes.}
          \label{fig:mask_edit}
        \end{figure}

\begin{figure}[htp]
  \centering
  \includegraphics[width=1\textwidth]{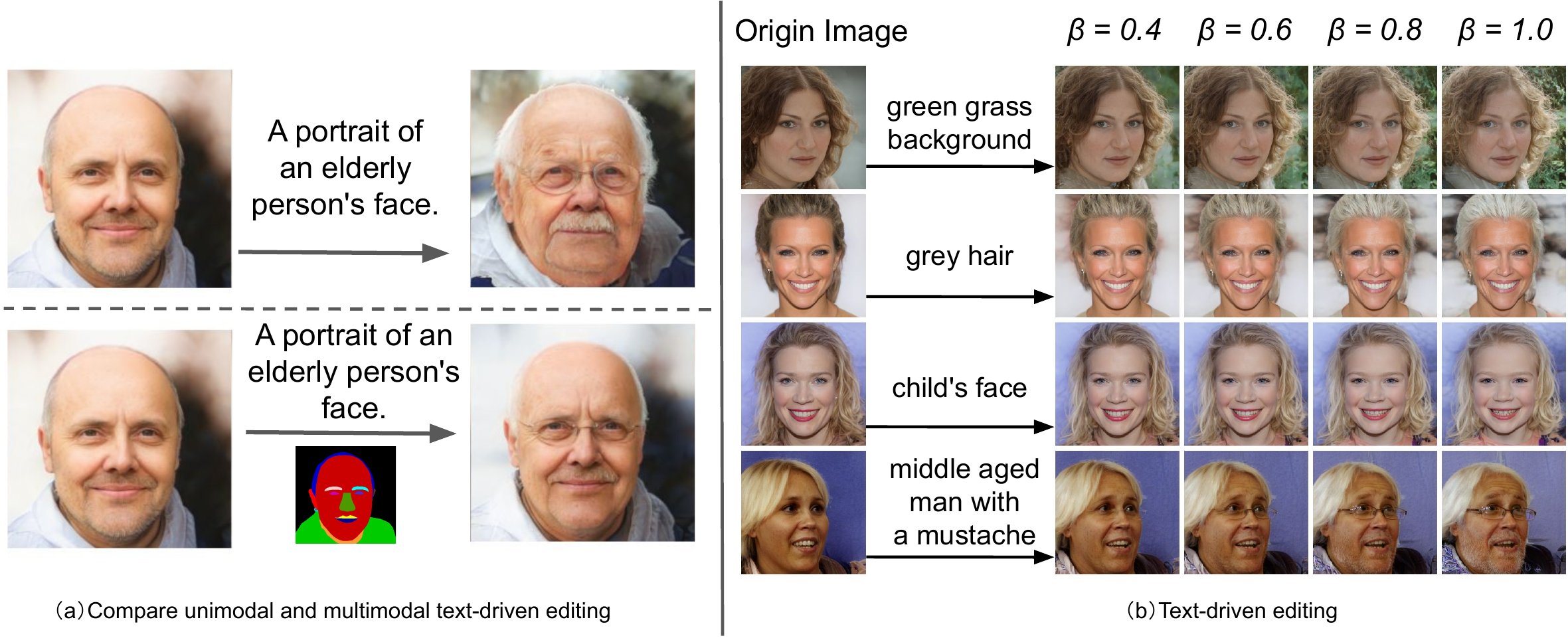}
  \caption{Text-driven image editing. (a) The multimodal text-driven editing in our framework shows more faithful results, effectively fixing the facial shape and avoiding unwanted changes, (b) real image editing with changed degree.}
  \label{fig:Generation_multimodal_editing_degree}
\end{figure}

\noindent \textbf{CMMD.} We employ CLIP Maximum Mean Discrepancy (CMMD) \cite{jayasumana2024rethinking} to measure the image realistic quality. Unlike the Fréchet Inception Distance (FID), which relies on Inception embeddings \cite{saritacs2024analyzing} and assumes normality in feature distributions, CMMD utilizes CLIP embeddings and Maximum Mean Discrepancy (MMD) distance. Inception embeddings, trained on ImageNet, primarily focus on general object recognition (e.g., animals, products) and are less effective for facial feature extraction. In contrast, CLIP \cite{Radford2021LearningTV}, trained on a dataset 400 times larger than ImageNet, demonstrates superior performance in evaluating facial data \cite{saritacs2024analyzing} and capturing facial attributes \cite{patashnik2021styleclip,Pinkney2022clip2latentTD}. Furthermore, the MMD metric of CMMD does not impose distributional assumptions like FID. Therefore, we use CMMD to assess the generated realistic quality of facial images.

\begin{figure*}[htp]
  \centering
  \includegraphics[width=1\textwidth]{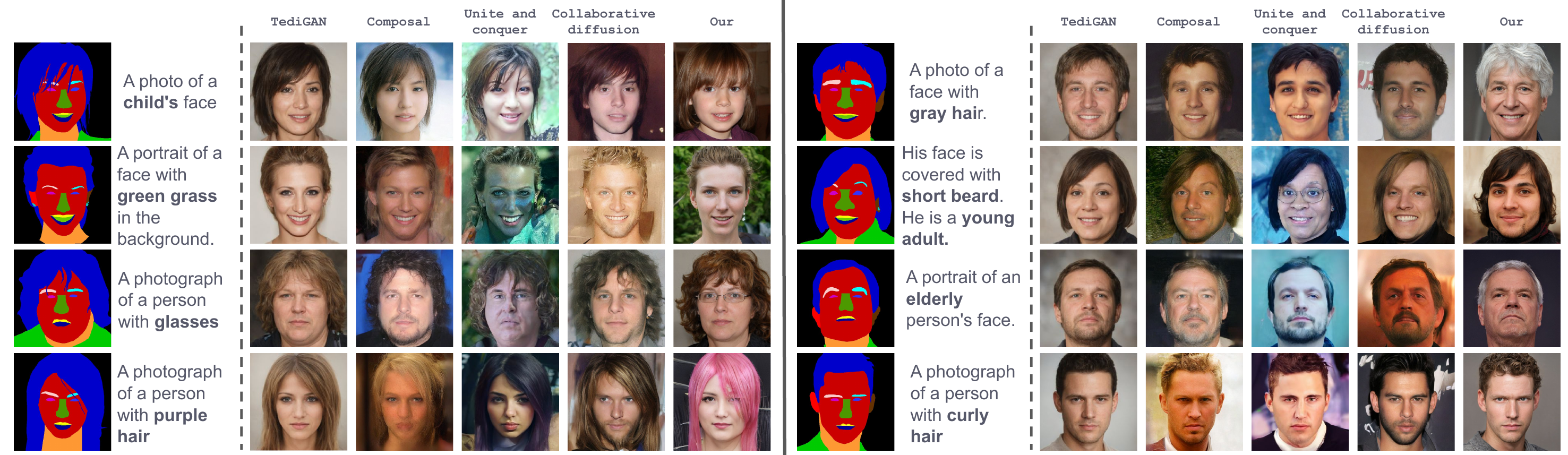}
  \caption{Image Generation compared with baseline. The left part of the dotted line is the multimodal conditional input. And the right part it the generated images.}
  \label{fig:Generation_multimodal_baseline}
\end{figure*}

\begin{table}
   \centering
    \caption{Comparison with the state-of-the-art Multimodal2Face generation method. The higher performance is better. \textbf{Text (\%) $\uparrow$} indicate CLIP Score.}      
    \small
    \begin{tabular}{l c c c}
    \midrule
    \textbf{Method} & \textbf{Text (\%) $\uparrow$} &\textbf{Mask (\%) $\uparrow$} & \textbf{CMMD $\downarrow$} \\

    \midrule
    TediGAN ~\cite{xia2021tedigan} & 22.53 & 82.86 & 1.70\\
    Composable ~\cite{Liu2022CompositionalVG} & 23.52& 80.76 & 2.55 \\
    UniteConquer ~\cite{nair2023unite} &  23.22& 85.29 & 1.53\\
    Collaborative Diffusion ~\cite{huang2023collaborative} & 23.48 & 82.96 & 1.98 \\

    \midrule
    MM2Latent &\textbf{\textcolor{red}{24.59}} & \textbf{\textcolor{red}{85.61}} & \textbf{\textcolor{red}{1.43}}  \\
    \midrule
  \end{tabular}
  \label{tab:multimodal}
\end{table}

\subsection{Quantitative Analysis}
\textbf{Comparison with the state-of-the-arts (SoTA).}
We compared our proposed~MM2Latent with recent advancements in multimodal text-to-image generation on the CelebA-Dialog/Mask dataset, with results detailed in Table~\ref{tab:multimodal}. From this table, it is evident that our~MM2Latent~achieves SoTA performance in terms of text consistency and mask accuracy, showing notable improvements over competing methods. Notably, our method is the only one achieving a CLIP score higher than $24\%$. While UniteConquer closely matches our method in mask accuracy, it significantly lags in text consistency. Furthermore, our method also registers the lowest CMMD score, indicating superior image quality. Given our leading performance in mask accuracy, text consistency, and CMMD distance, the results clearly demonstrate the effectiveness of our approach.

\noindent \textbf{Why MM2Latent works better.} There are two reasons: 1) learnable multimodal fusion, and 2) the Pseudo Text Embedding Generation (PTEG) improves inference robustness. Unlike TediGAN, Compositional, and UniteConquer, which rely on manual feature fusion consequently yield sub-optimal performance, our method uses end-to-end learnable feature fusion, making it easier to optimize. While Collaborative Diffusion also uses learnable multimodal fusion, it overlooks inference robustness — In training, text and masks are paired from dataset, but in inference, the mask might be combined with various text prompts (e.g., the user may generate people with different attributes using the same mask). Our PTEG module addresses this by generating multiple pseudo text embeddings from a single sample pair. This approach simulates inference situation and enhances robustness during the inference process.

\noindent \textbf{Ablation study of~MM2Latent}. We conducted ablation experiments to evaluate the PTEG module, batch normalization (BN) layers, and dropout layers, as detailed in Table~\ref{tab:ablation_componets}. These experiments utilized a MappingNetwork based on 8-layer fully-connected layers. From the results: Pseudo-text feature generation slightly reduced text consistency but significantly improved mask accuracy and substantially lowered the CMMD distance. Overall, the inclusion of this module enhanced performance. Dropout layers introduced noise that complicated the learning process, leading to a decrease in performance metrics. Batch normalization (BN) adjusted the distribution of input modalities, simplifying the learning process and mitigating difficulties introduced by the dropout layer. The combination of these three modules achieved a balanced performance, yielding the most effective results. While text consistency was marginally lower than the baseline, mask accuracy saw a significant increase. Although CMMD was slightly higher, it still demonstrated good image quality compared to the state-of-the-art results in Table~\ref{tab:multimodal}. Thus, this strategic integration of modules adopted for multimodal design in our experiments.

\begin{table}
\centering
    \small
    \caption{Ablation study of different componets. \Checkmark means involve this componets in the framework. These experiments are evaluated on our 8 layer MappingNet.}   
    \begin{tabular}{c c c c c c}
    \midrule
    \textbf{Pseudo Text Embedding} & \textbf{BN} & \textbf{Drop} & \textbf{Text (\%) $\uparrow$} &\textbf{Mask (\%) $\uparrow$} & \textbf{CMMD $\downarrow$} \\

    \midrule
    - & - & - &  \textbf{\textcolor{red}{24.46}} & 82.50 & 1.43\\
    \Checkmark & - & - & 24.37 & 84.53 & \textbf{\textcolor{red}{1.23}}\\
    \Checkmark & \Checkmark & - & 24.39 & 84.60 & 1.24 \\
    \Checkmark & - & \Checkmark & 24.22 & 71.43 & 2.06\\
    \Checkmark & \Checkmark & \Checkmark & \textbf{\textcolor{red}{24.43}} & \textbf{\textcolor{red}{85.17}} & 1.40\\

    \midrule
  \end{tabular}
  \label{tab:ablation_componets}
\end{table}

We also conducted an ablation study focusing on the number of layers in the MappingNetwork, with findings presented in Table~\ref{tab:ablation_num_layers}. The results clearly indicated that increasing the number of layers generally enhances performance. Based on these observations, we adopted a configuration of 12 layers for our final experiments, both for quantitative and qualitative comparisons with SoTA methods in table~\ref{tab:multimodal}. We argue that incorporating even more layers could potentially further enhance the performance of our methods.

\begin{table}
    \centering
    \small
    \caption{Ablation study of different number of FC layers. The setting 12 layers are used to compare with SoTA methods.}       
    \begin{tabular}{c c c c}
    \midrule
    \textbf{Number of Layers} & \textbf{Text (\%) $\uparrow$} &\textbf{Mask (\%) $\uparrow$} & \textbf{CMMD $\downarrow$}\\ 
    \midrule
    4 & 24.29 & 84.57 & 1.47 \\
    8 & 24.43 & 85.17 & \textbf{\textcolor{red}{1.40}}\\
    12 & \textbf{\textcolor{red}{24.59}} & \textbf{\textcolor{red}{85.61}} & \textbf{\textcolor{red}{1.43}}\\

    \midrule
  \end{tabular}
  \label{tab:ablation_num_layers}
\end{table}

\noindent \textbf{Inference Speed.}
Real-time performance is crucial in image generation, as high memory costs and time consumption can restrict the practical applicability of a method. Although recent diffusion generative models offer many benefits, they suffer from significantly slower inference speeds compared to GANs. As illustrated in Table~\ref{tab:inference_speed}, our method not only maintains the best generation performance but also achieves the fastest inference speed. Compared with the leading diffusion-based model, our method is substantially quicker—almost 150 to 1300 times faster. Also, it is 4.82 times faster than TediGAN. 

    \begin{table}
    \centering
    \small
    \caption{Conduct inference speed tests on P100 GPUs, and provide the average results based on 100 inference runs.}       
    \begin{tabular}{c c c}
        \midrule
        \textbf{Method} & Generation Model  & Speed (ms)$\downarrow$\\
        \midrule
            Composable              &  Diffusion & 6,300.56 \\
            Unite and conquer       & Diffusion & 57,214.14 \\
            Collaborative Diffusion & Diffusion & 11,071.77\\
            TediGAN                 &  GANs & 114.02 \\
        \midrule
        MM2Latent & GANs & \textbf{\textcolor{red}{41.78}}\\
        \midrule
    \end{tabular}
    \label{tab:inference_speed}
    \end{table}

\subsection{Qualitatives Analysis}
\subsubsection{Multimodal image generation.} In Fig.~\ref{fig:Generation_multimodal_baseline}, we present a quality comparison of the generation results from our method against baseline methods on diverse attributes such as age, background, glasses, hair color, beard, and hairstyle. It is evident that our method generates realistic outputs that are consistent with the multimodal conditions. Our approach produces more plausible face images with high consistency between image-text and image-mask. The text and mask modalities exhibit excellent complementarity: the mask delineates the shape, outline of the generated human, while the text specifies attributes that the mask alone cannot convey, such as age, hair color and beard presence. For additional image generation results, see Fig.~\ref{fig:mask_sketch_3dmm_synthetic}.

\subsubsection{Multimodal real image editing.} In Fig.~\ref{fig:Generation_multimodal_editing_degree} and~\ref{fig:mask_edit}, we demonstrate real image editing, including text, mask, sketch, and 3DMM editing. The results shows our method's exceptional editing quality. For text-driven editing, we highlight the ability to adjust the editing strength at different scales, enabling precise control over attributes such as hair color and age through the parameter $\beta$. For mask, sketch, and 3DMM-driven editing, given their specific spatial requirements, we utilize the default setting of $\beta=1$ without need to modify the scale. This standard setting consistently delivers stable quality across all editing types, showcasing the robustness and versatility of our approach in diverse editing scenarios.

\section{Conclusions}\label{sec:conclusions}
    
    Our research contributes to multimodal image generation, which explores the advantages and complements of various modalities to achieve more control and innovative image synthesis. For instance, we can utilize the advantages of text in controlling diverse attributes and masks in controlling spatial locations. In our work, we aim to utilize text, spatial mask, sketch, and 3DMM modalities. Previous SoTA methods in this field are limited by their requirement for many hyperparameters in the inference stage, rely on manual operations, have significant computational demands both in training and inference or inability to edit real images. We addresses these issues by introducing~MM2Latent, a novel framework based on StyleGAN2. The method achieves SoTA results in multimodal consistency and image realistic quality while also having the fastest inference speed. Also, it demonstrates realistic results in multimodal image editing.


{\setlength{\parindent}{0cm}
\textbf{Acknowledgments:} This work was supported by the EU H2020 AI4Media No. 951911 project.
}

\clearpage



%
%
\bibliographystyle{splncs04}
\bibliography{main}
\end{document}